# FCMI: Feature Correlation based Missing Data Imputation

Prateek Mishra [1], Kumar Divya Mani [2], Prashant Johri [3], Diksha Arya [4]

[1,2,3,4] Department of Computer Science, Galgotias University, India
New Delhi, India
`pr4t333k@gmail.com`

**Abstract.** Processed data are insightful, and crude data are obtuse. A serious threat to data reliability is missing values. Such data leads to inaccurate analysis and wrong predictions. We propose an efficient technique to impute the missing value in the dataset based on correlation called FCMI (Feature Correlation based Missing Data Imputation). We have considered the correlation of the attributes of the dataset, and that is our central idea. Our proposed algorithm picks the highly correlated attributes of the dataset and uses these attributes to build a regression model whose parameters are optimized such that the correlation of the dataset is maintained. Experiments conducted on both classification and regression datasets show that the proposed imputation technique outperforms existing imputation algorithms.

## 1   Introduction

Pre-processed data having prevalent missing values is a quintessential challenge encountered in machine learning and data analysis [1,2]. We amass nearly 2.5 quintillion bytes of data each day[13]. Much of these consist of missing values and affect the supervised learning model [2,3]. The missing values can affect the result of the analysis. That's why they must be detected and imputed. Unfortunately, there doesn't exist a panacea of imputation technique. There are lots of approaches to deal with missing values but no one captures the result appropriately. Available imputation techniques do not perform well and often neglect the correlation between attributes of the dataset and the column having missing values.

In our approach, we have kept correlation as the central theme. We acknowledge that a highly correlated column reflects a relationship and interdependency (in most cases). As an illustrative example, we can assume two columns (it can be more columns as well). If the correlation between them is high, we say that they capture information about each other. If one column happens to have a missing value in one of its instances, we can use the other column to predict the missing value. That is our strategy. Correlation is not causation, and we have thoroughly followed this ideology. We developed a loss function to strike a balance so that a highly correlated column may not generate redundancy. Our loss function makes sure that the issue of multicollinearity[15] does not arise. We have explained our algorithms in detail in the later parts of this paper.

The remainder of the paper is organized as follows: In section 2, we have discussed the related works on missing data imputation. Then in "Preparation and Methodology" Section 3, we have explained the dataset preparation strategy and our proposed imputation algorithm FCMI. In section 4, we have discussed the performance of our algorithm. And then, Section 5 presents a conclusion of our paper with emphasis on future works.

## 2  Related Works

In the area of data analysis, in recent years, a lot of emphasis has been given to the development of imputing missing values in a dataset. The existing ones do not capture the principle of correlation. In this section, an emphasis has been given on the popular imputation techniques.

Zhang et al. [6], in their paper, presented an imputation method based on clustering. They adopted the kernel method to fill in the missing value of a particular instance with the reliable value generated from the available instances. This approach is sophisticated. In practical scenarios, datasets generally have missing values in both class and conditional attributes.

Melissa et al.[4], in their paper on MICE (Multiple Imputation by Chained Equations) worked under certain assumptions about the missingness mechanism (MAR, MCAR). Multiple copies of the original datasets get created, and the replacement of missing values happens using a statistical approach. The results of these analyses are pooled and reported. It imputes the missing values in a dataset by focusing on one value at a time. All the available values are used to predict a particular missing value. It works on the regression model technique and becomes impractical to use for massive datasets because of its complexity. The other disadvantage is the lack of theoretical justification which is present in other imputation techniques.

Khan and Hoque [5], in their paper, presented an approach that was a mixture of single and multiple imputation techniques for replacing missing data. Their approach is an extension of MICE and deals with both categorical and numerical data But, it fails to capture the essence of correlation and thus limits the scope of accurate prediction of missing values.

## 3  Preparation and Methodology

### 3.1  Dataset Preparation

For our experiments, we have used five datasets collected from Kaggle Website and the UCI Machine Learning repository. We have also introduced 10% missing values in all the datasets based on the directions of [7]. The categorical attributes present in the datasets are transformed into numerical attributes using Label Encoding and One-Hot encoding techniques. The details of the datasets used are tabulated in Table 1.

**Table 1:** Details of the datasets used

| No. | Dataset | Attributes | Rows | Missing Values | Type |
|---|---|---|---|---|---|
| 1 | Car Evaluation Dataset [8] | 6 | 1728 | 173 | Classification |
| 2 | House Pricing Dataset [9] | 81 | 1460 | 146 | Regression |
| 3 | Forest Fire Dataset [10] | 13 | 517 | 52 | Regression |
| 4 | Wine Quality Dataset [11] | 12 | 4898 | 490 | Classification |
| 5 | IRIS Dataset [12] | 4 | 150 | 15 | Classification |

### 3.2 Evaluation criteria

The evaluation criteria used for the comparison of various imputation algorithms depends on the type of the dataset.

The Root Mean Square Error (RMSE) metric is used for the evaluation of regression datasets. It computes the difference between the observed value and the imputed value (predicted value). It is defined as :

$$RMSE = \sqrt{1/n \sum_{i=1}^{n}(y_i - y_i^*)^2}$$

For n = 5, let's assume the predicted values be $y^*_i$ = [1.6,4.5,6.1,7.9,10.1] and the true values be $y_i$ = [2,4,6,8,10] then, RMSE score will be 0.2966 .

Accuracy metric is used for the evaluation of classification datasets. It's defined as the ratio of True prediction to the Total predictions by the algorithm.

$$ACCURACY = \frac{TP + TN}{TP + FP + TN + FN}$$

For n = 5, let's assume the predicted values be [1,0,1,1] and the actual values be [1,1,0,1] then classification accuracy will be 0.5 .

### 3.3 Proposed Algorithm

Our algorithm FCMI (Feature Correlation based Missing Data Imputation) has been introduced in this section. The missing values of an attribute are imputed by our algorithm by developing a linear regression model. The Predictors are the top three ( K = 3 ) attributes of the dataset whose correlation coefficient with the target column (column having missing values) is maximum. We have also described the complete FCMI algorithm in Algorithm 1.

**Algorithm 1: Feature Correlation based Missing data Imputation**

    **Input:** x: columns with missing values
    m: dataset with no missing values
    n: dataset containing only missing values
    **Output:** m: updated dataset with missing values imputed.

1. **for** *each missing column (i) in x* **do :**
2.     **for** *each column (j) in the dataset* **do :**
           Calculate *correlation coefficient* between ( **m[i] and m[j]** ) and store it in **P**
       **end**
3.     **Set** 'K' = top 3 columns names whose correlation is maximum w.r.t **m[i]**
4.     **Build** *a linear regression model* and train it on predictor variables **k1, k2, k3** and target as **m[i]**. [The parameters of the linear regression model is optimized using loss function described in Algorithm 2 ]
5.     **Predict** the values of **n[i]** using the model trained in 4.
6. **end**

The linear regression model trained for missing data imputation is optimized using a loss function containing two components. The first component of the loss function is dependent on the type of the target variable. If the target column is continuous, the first part is Squared Error Loss. And, it will be Categorical Cross Entropy loss for the categorical values in the target column.

The second component of the loss function is KL divergence loss. It is evaluated using a vector of the initial correlation coefficient and the vector containing the correlation coefficient of the predicted target variable. We have mentioned the loss function in equations (1) and (2). We have also stated it in Algorithm 2.

$$\text{Loss} = (y - f(x))^2 + KL(P \| Q) \tag{1}$$

$$\text{Loss} = (1-y)\log(1-p)) - (y\log(p) + KL(P \| Q) \tag{2}$$

The objective behind the training is to make sure that the trained model learns to predict missing values that do not disturb the existing correlation between the columns of the dataset. The proposed loss function contains KL divergence loss which measures how the two vectors containing correlation coefficient calculated initially and after prediction is different. The KL divergence value is 0 if the two vectors are exactly the same. When the correlation is maintained by the model then the value of the resultant loss function will only depend on the difference between the observed and predicted values

---

**Algorithm 2: Loss function for FCMI**

    **Input:**   x: true values of the column

    y: predicted values of the column

    z: dataset with the only **k1, k2, k3** attributes

    P: initial vector containing correlation coefficients.

    **Output:** r: calculated loss.

1. **Calculate: if** *x is Continuous* **then,**

              **E** = *squared error loss on **x** and **y**.*

       **else**

               **E** = cross entropy loss between x and y.

2. **Calculate : Q =** *a vector with correlation coefficient* between  y **and** z[K1],

        y **and** z[k2], y and z[k3].

3. **Calculate : Loss = E + KL( P || Q)**

4. **Return Loss**

---

## 3.4     Experimental Design

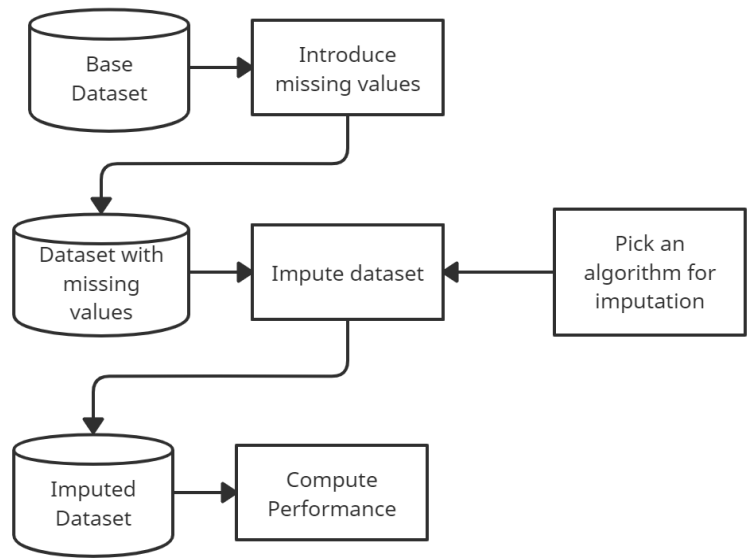

**Figure 1:** Block diagram representing the Experimental Design of the system

We have presented our experimental design in figure 1. At the start, we have a dataset with no missing values. We refer to it as the base dataset. Then, we introduce 10% missing values to some of the columns of the dataset chosen randomly except the target column. After adding missing values, we pick up an imputation algorithm for imputing the missing values present in the new dataset. Once the missing values are imputed, the performance of the dataset is then imputed using two different metrics i.e. RMSE or Accuracy metric based on the type of dataset.

## 4   Results and Discussion

For evaluating the performance of our FCMI algorithm we implement three other imputation algorithms and then compare the accuracy with our algorithm. The algorithms used for missing data imputation are MICE, SICE, KNN impute, and FCMI. In our experiments, we have used two types of datasets. One contains the continuous target variable and the other contains the categorical target variable. We have compared our algorithms separately on those two types of datasets and presented them in Table 2 and Table 3. Our experiments showed that the accuracy of KNN imputation was the lowest for both classification and regression datasets. The accuracy of the MICE imputation algorithm was better than the KNN imputation every time and it performs better than the SICE algorithm when the size of the dataset is small or when the target variable is continuous.

**Table 2:** Accuracy score of imputation algorithms on various classification datasets.

| Algorithm | Car Evaluation Dataset | Wine Quality Dataset | IRIS Dataset |
|---|---|---|---|
| MICE | 65.8 | 71.2 | 81.3 |
| SICE | 74.3 | 77.4 | 81.5 |
| KNN | 61.2 | 68.4 | 78.3 |
| **FCMI** ( **Proposed Algorithm** ) | **78.1** | **78.3** | **81.68** |

On using the FCMI algorithm for imputation, we found that it outperformed both regression and classification datasets. The RMSE score of the FCMI algorithm was much higher than other algorithms because the correlation (High or Low ) between the continuous columns is quite visible than by categorical columns and the FCMI algorithm captures this correlation information while training the model very efficiently. The expressiveness of the continuous columns is the reason behind the high RMSE score achieved by the FCMI algorithm. When the FCMI algorithm was used upon small-size datasets we found the performance to be comparatively similar to other existing algorithms. The correlation between columns when the sample size is small is highly unreliable because of the increased variability and lower probability of repetition of values [14]. When the correlation between the columns is not reliable then, the performance of the FCMI algorithm also gets affected. The influence of this unreliability is balanced by the first part of the loss function which measures the difference between predicted and observed values.

**Table 3:** RMSE score of imputation algorithms on various regression datasets.

| Algorithm | Housing Price Dataset | Forest Fire Dataset |
|---|---|---|
| MICE | 90.3 | 93.4 |
| SICE | 89..4 | 92.1 |
| KNN | 82.1 | 87.3 |
| **FCMI** ( **Proposed Algorithm** ) | **92..3** | **93.8** |

## 5 Conclusion

In this paper, we propose a novel, simple, and effective algorithm for the imputation of missing values assuming values are MAR (Missing at Random). The principle idea behind the proposed algorithm is to pick the highly correlated attributes of the dataset w.r.t column with missing values and use these features to build a regression model whose parameters are optimized such that the correlation of the dataset is maintained. The results of our experiments prove the assumption that a pair of columns that are highly correlated have captured the information about each other and thus can be used for imputation if any of their instances are missing. On comparing the performance of FCMI we found that it performed better than the existing algorithms. Despite, its good performance in most of the cases it still does not predict the missing outliers of the dataset accurately. We believe that our principal idea about a correlation can be used for the improvement of other existing algorithms and it can also be further investigated for improved prediction of outliers.